\pdfoutput=1
\documentclass[11pt]{article}

\usepackage[]{emnlp2021}

\usepackage{times}
\usepackage{latexsym}

\usepackage[T1]{fontenc}
\usepackage[utf8]{inputenc}

\usepackage{microtype}

\usepackage{booktabs}
\usepackage{multirow}
\usepackage{array}
\usepackage{amsmath}
\usepackage{adjustbox}
\usepackage{url}
\usepackage{bm}
\usepackage{graphicx} 
\usepackage{natbib}
\usepackage{caption}
\usepackage{amssymb}

\newcommand{\tname}{ZeXT }
\newcommand{\tnamec}{ZeXT}
\newcommand{\mname}{SixT }
\newcommand{\mnamec}{SixT}
\usepackage{xcolor}
\newcommand{\tub}[1]{\underline{\textbf{#1}}}
\newcommand{\mtrv}{\multirow{5}{*}{\rotatebox{90}{Vanilla}}}
\newcommand{\mtrs}{\multirow{5}{*}{\rotatebox{90}{SixT}}}

\usepackage{listings}
\definecolor{codegreen}{rgb}{0,0.6,0}
\definecolor{codegray}{rgb}{0.5,0.5,0.5}
\definecolor{codepurple}{rgb}{0.58,0,0.82}
\definecolor{backcolour}{rgb}{0.95,0.95,0.92}

\lstdefinestyle{mystyle}{
    backgroundcolor=\color{backcolour},   
    commentstyle=\color{codegreen},
    keywordstyle=\color{magenta},
    numberstyle=\tiny\color{codegray},
    stringstyle=\color{codepurple},
    basicstyle=\ttfamily\footnotesize,
    breakatwhitespace=false,         
    breaklines=true,                 
    captionpos=b,                    
    keepspaces=true,                 
    numbers=left,                    
    numbersep=5pt,                  
    showspaces=false,                
    showstringspaces=false,
    showtabs=false,                  
    tabsize=2
}

\title{Zero-Shot Cross-Lingual Transfer of Neural Machine Translation \\ with Multilingual Pretrained Encoders}

\author{
  Guanhua Chen$^{1}$\thanks{\ \ \ Contribution during internship at Microsoft Research.},
  Shuming Ma$^{2}$,
  Yun Chen$^{3}$\thanks{\ \ \ Corresponding author.},
  Li Dong$^{2}$ \\
  \textbf{
  Dongdong Zhang$^{2}$,
  Jia Pan$^{1}$, 
  Wenping Wang$^{4,1}$,  
  Furu Wei$^{2}$ } \\
  $^1$The University of Hong Kong; 
  $^2$Microsoft Research \\
  $^3$Shanghai University of Finance and Economics;
  $^4$Texas A\&M University \\
  \{ghchen,jpan,wenping\}@cs.hku.hk, yunchen@sufe.edu.cn, \\ 
  \{shumma, lidong1, dozhang, fuwei\}@microsoft.com
}
\date{}

\begin{document}
\maketitle

\begin{abstract}
Previous work mainly focuses on improving cross-lingual transfer for NLU tasks with a multilingual pretrained encoder (MPE), or improving the performance on supervised machine translation with BERT. However, it is under-explored that whether the MPE can help to facilitate the cross-lingual transferability of NMT model. In this paper, we focus on a zero-shot cross-lingual transfer task in NMT. In this task, the NMT model is trained with parallel dataset of only one language pair and an \textit{off-the-shelf} MPE, then it is directly tested on zero-shot language pairs. We propose SixT, a simple yet effective model for this task. SixT leverages the MPE with a two-stage training schedule and gets further improvement with a position disentangled encoder and a capacity-enhanced decoder. Using this method, SixT significantly outperforms mBART, a pretrained multilingual encoder-decoder model explicitly designed for NMT, with an average improvement of 7.1 BLEU on zero-shot any-to-English test sets across 14 source languages. Furthermore, with much less training computation cost and training data, our model achieves better performance on 15 any-to-English test sets than CRISS and m2m-100, two strong multilingual NMT baselines. 

\end{abstract}

\section{Introduction}
Multilingual pretrained encoders~(MPE) such as mBERT~\cite{mbert}, XLM~\cite{conneau2019cross}, and XLM-R~\cite{xlmr} have shown remarkably strong results on zero-shot cross-lingual transfer mainly for natural language understanding (NLU) tasks, including named entity recognition (NER), question answering (QA) and natural language inference (NLI). These methods jointly train a Transformer \cite{transformer} encoder to perform masked language modeling task in multiple languages. The pretrained model is then fine-tuned on a downstream NLU task using labeled data in a single language and evaluated on the same task in other languages. With this pretraining and fine-tuning approach, the MPE is able to generalize to other languages that even do not have labeled data.
\begin{figure}[t]
  \center
  \includegraphics[width=0.9\columnwidth]{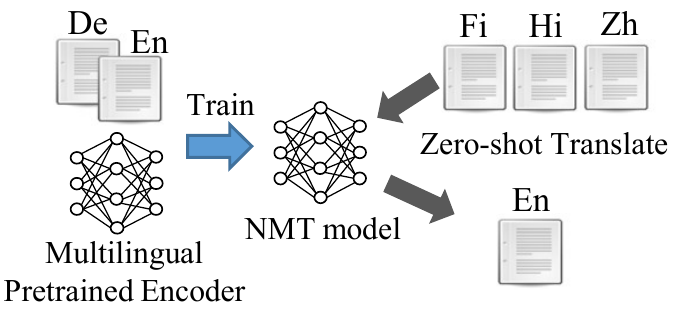}
  \caption{In the zero-shot cross-lingual NMT transfer task, the model is trained with parallel dataset of only one language pair (such as De-En) and a multilingual pretrained encoder. The trained model is tested on many-to-one language pairs (like Fi/Hi/Zh-En) in a zero-shot manner. Monolingual text of the to-be-tested source languages is not available in this task.}
  \label{fig:task-example}
\end{figure}
Given that MPE has achieved great success in cross-lingual NLU tasks, a question worthy of research is how to perform zero-shot cross-lingual transfer in the NMT task by leveraging the MPE. Some work~\cite{zhu20incorporating,yang20towards,weng20acquiring,imamura2019recycling} explores approaches to improve NMT performance by incorporating monolingual pretrained Transformer encoder such as BERT~\cite{bert}. However, simply replacing the monolingual pretrained encoder in previous studies with MPE does not work well for cross-lingual transfer of NMT (see baselines in Table~\ref{tab:task-results}). Others propose to fine-tune the encoder-decoder-based multilingual pretrained model for cross-lingual transfer of NMT \cite{mbart,lin2020pre}. It is still unclear how to conduct cross-lingual transfer for NMT model with existing multilingual pretrained encoders such as XLM-R. 

In this paper, we focus on a \underline{Ze}ro-shot cross-lingual(\underline{X}) NMT \underline{T}ransfer task (\tnamec, see Figure~\ref{fig:task-example}), which aims at translating multiple unseen languages by leveraging an MPE. Different from unsupervised or multilingual NMT, only an MPE and parallel dataset of one language pair such as German-English are available in this task. The trained model is directly tested on many-to-one test sets in a zero-shot manner.

We propose a \underline{Si}mple cross-lingual(\underline{X}) \underline{T}ransfer NMT model (\mnamec) which can directly translates languages unseen during supervised training. We initialize the encoder and decoder embeddings of \mname with the XLM-R and propose a two-stage training schedule that trades off between supervised performance and transferability. At the first stage, we only train the decoder layers, while at the second stage, all model parameters are jointly optimized except the encoder embedding. We further improve the model by introducing a position disentangled encoder and a capacity-enhanced decoder. The position disentangled encoder enhances cross-lingual transferability by removing residual connection in one of the encoder layers and making the encoder outputs more language-agnostic. The capacity-enhanced decoder leverages a bigger decoder than vanilla Transformer base model to fully utilize the labelled dataset. Although trained with only one language pair, the \mname model alleviates the effect of `catastrophic forgetting' \cite{serra18overcoming} and can be transferred to unseen languages. \mname significantly outperforms mBART with an average improvement of 7.1 BLEU on zero-shot any-to-English translation across 14 source languages. Furthermore, with much less training computation cost and training data, the \mname model gets better performance on 15 any-to-English test sets than CRISS and m2m-100, two strong multilingual NMT baselines.\footnote{The code is available at \url{https://github.com/ghchen18/emnlp2021-sixt}.}  

\section{Problem Statement} \label{sec:taskdef}
The zero-shot cross-lingual NMT transfer task (\tnamec) explores approaches to enhance the cross-lingual transferability of NMT model. Given an MPE and parallel dataset of a language pair $l_s$-to-$l_t$, where $l_s$ and $l_t$ are supported by the MPE, we aim to train an NMT model that can be transferred to multiple unseen language pairs $l^i_z$-to-$l_t$, where $l^i_z \neq l_s$ and $l^i_z$ is supported by the MPE. The learned NMT model is directly tested between the unseen language pairs $l^i_z$-to-$l_t$ in a zero-shot manner. Different from multilingual NMT \cite{johnson2017google}, unsupervised NMT \cite{lample2018phrase} or zero-resource NMT through pivoting \cite{chen2017teacher,chen2018zero}, neither the parallel nor monolingual data in the language $l^i_z$ is directly accessible in the \tname task. The model has to rely on the off-the-shelf MPE to translate from language $l^i_z$. The challenge to this task is how to leverage an MPE for machine translation while preserving its cross-lingual transferability. In this paper, we utilize XLM-R, which is jointly trained on $100$ languages, as the off-the-shelf MPE.
\begin{table*}[t]
  \centering 
  \resizebox{1.38\columnwidth}{!}{
    \begin{tabular}{llrrrrr}
    \toprule
    ID  & Strategy                    & Es    & Fi    &  Hi  & Zh   & Avg. \\ \midrule
    (1) & Vanilla Transformer (BaseDec) & 0.6   &  0.5  &  0.1 & 0.2  &  0.4  \\ \midrule
    \multicolumn{7}{l}{Encoder embedding:}  \\ 
    (2) & (1) + FT encoder embed      & 2.5   & 1.8   &  1.0 & 1.3  & 1.65    \\ 
    (3) & (1) + Fix encoder embed     & 2.4   & 1.5   &  1.3 & 1.7  & 1.73    \\  \midrule 
    \multicolumn{7}{l}{Encoder layers:} \\ 
    (4) & (3) + FT encoder layers     & 11.6  & 7.4   &  5.9 & 4.3  & 7.3    \\ 
    (5) & (3) + Fix encoder layers    & 17.9  & 10.1  &  6.0 & 5.2  & 9.8    \\  \midrule
    \multicolumn{7}{l}{Decoder embedding:}  \\
    (6) & (5) + FT decoder embed      & 18.7  & 10.3  &  7.1 & 6.3  & 10.6   \\ 
    (7) & (5) + Fix decoder embed     & 20.2  & 12.3  &  \tub{7.8}  & 6.3  &  11.6  \\   \midrule  
    \multicolumn{7}{l}{Decoder layers:}  \\ 
    (8) & (7) + Fix decoder layers    & 2.1   & 2.2   & 0.8  & 1.0  & 1.5    \\
    (9) & (7) + FT decoder layers     & 20.0  & 12.3  & 7.3 &  \tub{6.9}  & 11.6    \\
    (10) & (7) + Rand BigDec          & \tub{20.7}  & \tub{13.7} &  7.7 &  6.8  &  \tub{12.2}  \\ 
    \bottomrule
    \end{tabular}}
  \caption{BLEU results of different initialization and fine-tuning strategies on zero-shot any-to-English language pairs. Starting from vanilla Transformer where all parts are randomly initialized (Strategy (1)), we initialize the encoder embedding (Strategy (2)-(3)), the encoder layers (Strategy (4)-(5)), the decoder embedding (Strategy (6)-(7)) and the decoder layers (Strategy (8)-(10)) with MPE sequentially. Each time we compare the strategy of `FT' and `Fix' which fine-tunes the corresponding module or keeps it fixed, respectively. Since Strategy (8)-(9) use a larger decoder than the rest ones due to decoder layer initialization, we add Strategy (10) whose decoder size is the same as Strategy (8)-(9) for fair comparison. The best BLEU is bold and underlined.}
  \label{tab:init-results}
\end{table*} 

The \tname task calls for approaches to efficiently build a many-to-one NMT model that can translate from $100$ languages supported by XLM-R with parallel dataset of only one language pair. The trained model could be useful for translating resource-poor languages. It can further extend to scenarios where datasets of more language pairs are available. In addition, while currently the cross-lingual transferability of different MPEs is mainly evaluated on cross-lingual NLU tasks,  the \tname task provides a new perspective for the evaluation, which can hopefully facilitate the research on MPEs.

\section{Approach}
\subsection{Initialization and Fine-tuning Strategy} \label{sec:empirical}
For downstream tasks like cross-lingual NLI/QA, only an output layer is added to the pretrained encoder at the fine-tuning stage. In contrast, an entire decoder is added on top of the MPE when the model is adapted to NMT task. The conventional strategy that fine-tunes all parameters reduces the cross-lingual transferability in the pretrained encoder due to the catastrophic forgetting effect. Therefore, we make \emph{an empirical exploration} on how to initialize and fine-tune the NMT model with an MPE. The NMT model can be divided into four parts in our method: encoder embedding, encoder layers, decoder embedding, and decoder layers. With an MPE, each part can be trained with one of the following methods, namely,

$\bullet$ \noindent \textbf{Rand}: randomly initialized and trained;

$\bullet$ \noindent \textbf{Fix}: initialized from the MPE and fixed;

$\bullet$ \noindent \textbf{FT}: initialized from the MPE and trained.

\begin{figure*}[t]
  \center
  \includegraphics[width=1.3\columnwidth]{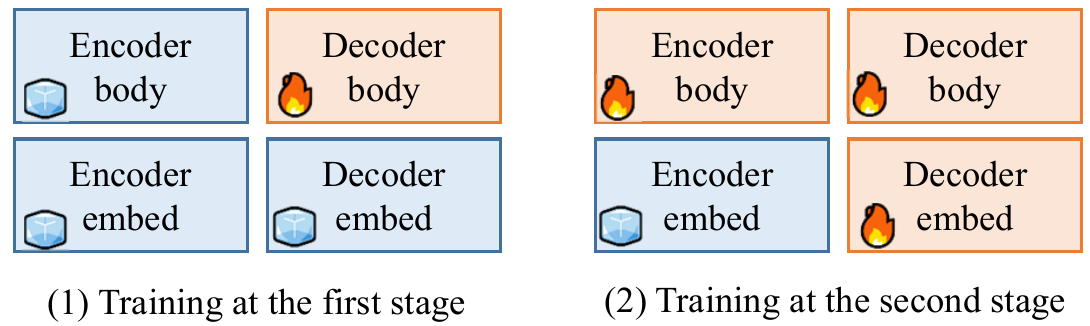}
  \caption{The best strategy for training NMT model for \tname task. The blue icy blocks are initialized with an MPE and frozen, while the red fiery blocks are initialized randomly or from the first stage.}
  \label{fig:model-example}
\end{figure*}

We compare different fine-tuning strategies for these modules in a greedy manner. Starting from vanilla Transformer where all parts are randomly initialized, we explore the best training method for the encoder embedding, the encoder layers, the decoder embedding, and the decoder layers, sequentially. The details of experimental settings are in the Section~\ref{sec:setup}. From the results shown in Table~\ref{tab:init-results}, we observe that it is the best to initialize the encoder embedding, the encoder layers and the decoder embedding with XLM-R and keep their parameters frozen, while randomly initializing the decoder layers (see Figure~\ref{fig:model-example}). More discussions are in the Section~\ref{sec:task-res}. 
\paragraph{Two-stage training} Since we freeze the encoder and only train the decoder layers, the model is able to perform translation while preserving the transferability of the encoder. However, freezing most of the parameters limits the capacity of the NMT model, especially when the training data goes large. Therefore, we propose a second training stage to further improve the translation performance by jointly fine-tuning all parameters except encoder embedding of the NMT.\footnote{According to our preliminary experiment, the average BLEU is 0.2 lower when the encoder embedding is also learned at the second stage. Besides, freezing encoder embedding leads to higher computational efficiency.} Since the decoder has been well adapted to the encoder at the first stage, we expect the model can be slightly fine-tuned to improve the translation capacity without losing the transferability of the encoder. 

\subsection{Model}
The training strategy and generalization objective of our model are different from vanilla Transformer. This motivates us to propose a new model that can further improve on zero-shot translations. The proposed model consists of a position disentangled encoder and a capacity-enhanced decoder, which aims at enhancing the cross-lingual transferability of the encoder and fully utilizing the labelled data, respectively.

\paragraph{Position disentangled encoder} The representations from XLM-R initialized encoder have a strong positional correspondence to the source sentence. The word order information inside is language-specific and may hinder the cross-lingual transfer from supervised source language to unseen languages. Inspired by \citet{liu2020improving}, we propose to relax this structural constraint and make the encoder outputs less position- and language-specific. More specifically, at the second stage, we remove the residual connection after the self-attention sublayer in one of the encoder layers $i$ during training and inference.\footnote{Different from \citet{liu2020improving}, we keep the layer normalization module after the self-attention sublayer for slightly better validation performance.} The other encoder layers remain the same. The hidden states in this $i^\mathrm{th}$ encoder layer are calculated as the following pseudo code:
\begin{lstlisting}[language=Python]
  h[i] = SelfAttn(h[i-1])
  h[i] = LayerNorm(h[i]) # No residual connection here
  h[i] = h[i] + LayerNorm(FFN(h[i]))
\end{lstlisting}
where \texttt{SelfAttn} is the encoder self-attention sublayer, \texttt{FFN} is the feed-forward sublayer and \texttt{LayerNorm} is the layer normalization. \citet{liu2020improving} aim at training a language-agnostic encoder for NMT using parallel corpus from scratch. Compared with them, our method shows that it's possible to make a pretrained multilingual encoder more language-agnostic by relaxing the position constraint during fine-tuning.

\paragraph{Capacity-enhanced decoder} Some previous work~\cite{zhu20incorporating,yang20towards} incorporates BERT into NMT and configures the decoder size as \citet{transformer}. For example, to train an NMT on Europarl De-En training dataset, the default decoder configuration is Transformer base~\cite{gu2018maml,currey2020distilling}. However, our model relies more on the decoder to learn from the labeled data, as the encoder is mainly responsible for cross-lingual transfer. This is also reflected in our training strategy: at the first stage only the decoder parameters are optimized, while at the second stage the encoder is only slightly fine-tuned to preserve its transferability. Therefore, the model capacity of \mname is smaller than vanilla Transformer with the same size. We propose to apply a capacity-enhanced decoder that has larger dimension of feed forward network, more layers and more attention heads at both the first and second training stages. The improvement brought by the big decoder is not simply because of more model parameters. More discussions are in the Section~\ref{sec:task-res}.

\section{Experiments} \label{sec:main-exp}
\subsection{Setup} \label{sec:setup}
\paragraph{Dataset} We focus on the any-to-English translations for the \tname task. The Europarl-v7 German and English is used as training set. We evaluate the cross-lingual transfer abilities of NMT models on a variety of languages from different language groups\footnote{We refer to the language group information in Table 1 of \citet{fan2020m2m}.}: German group (De, Nl), Romance group (Es, It, Ro), Uralic and Baltic group (Et, Fi, Lv), Indo-Aryan group (Hi, Ne) and Chinese (Zh). A concatenation of Fr-En and Cs-En validation dataset which are from different language groups is used as validation dataset for all any-to-English translation tasks. The details of the datasets are in the appendix. Note that none of the monolingual dataset of the tested source languages is available in all experiments.

\paragraph{Model settings} We use the XLM-R base model as the off-the-shelf MPE. The model is implemented on \texttt{fairseq} toolkit~\cite{ott2019fairseq}. We set Transformer encoder the same size as the XLM-R base model. For the decoder, we use the same hyper-parameter setting as the encoder. We denote model with such configuration as \mname and use this configuration for our NMT models through the paper unless otherwise stated. The encoder-decoder attention modules are randomly initialized. We remove the residual connection at the 11-th (penultimate) encoder layer, which is selected on the validation dataset. 

For the empirical exploration in Table~\ref{tab:init-results}, we use two model configurations. For Strategy (1)--(7) where decoder layers are trained from scratch, we use a smaller decoder denoted as BaseDec. This model configuration is denoted as \mname small. For the rest strategies, we follow the configuration of \mname and denote its decoder as BigDec. Table~\ref{tab:app-dec-config} in Appendix presents the details of different model configurations.  
\begin{table*}[t]
  \centering 
  \resizebox{1.0\textwidth}{!}{
    \begin{tabular}{l|r|rrrrrrrrrrr}
    \toprule
    Model                                      & De    & Nl    & Es   & It    & Ro    & Fi    & Lv   & Et     & Hi    & Ne    & Zh   & Avg. \\ \midrule
    Vanilla Transformer        & 22.6  & 0.6  &  0.6  & 0.2   & 0.7   & 0.5   &  0.2   & 0.4   & 0.1  & 0.1   & 0.2  & 0.4    \\ 
    \quad + XLM-R fine-tune encoder      & 17.9  & 22.9 & 14.1  & 15.9  & 13.5  & 7.9   &  6.6   & 6.6   & 5.7  & 4.5   & 5.6  & 10.3   \\ 
    \quad + XLM-R as encoder embedding   & 22.7  & 25.7 & 13.9  & 16.1  & 11.4  & 7.2   &  5.9   & 6.2   & 4.4  & 2.6   & 3.9  & 9.7    \\
    \quad + Recycle XLM-R for NMT        & 23.0 & 28.3  & 17.5  & 21.8  & 16.3  & 9.2   & 7.1    & 8.3   & 5.1  & 4.0   & 4.5  & 12.2    \\
    \quad + XLM-R fused model                    & 23.6  & 24.9 & 10.7  & 10.0  & 10.5  & 6.1   &  4.0   & 5.1   & 6.0  & 3.4   & 6.1  & 8.7    \\
    \quad + XLM-R fine-tune all          & 20.2  & 28.3 & 17.2  & 21.4  & 17.2  & 11.3  &  8.4   & 8.7   & 6.1  & 5.0   & 5.8  & 12.9  \\ 
    \midrule                        
    \textit{Our proposed \mnamec}              &  \tub{26.4}  &  \tub{38.6}  & \tub{22.9} & \tub{32.0} & \tub{23.9} & \tub{15.8} &  \tub{12.0}  & \tub{14.5} & \tub{8.6} &  \tub{6.1} &  \tub{8.1}  &  \tub{18.3}    \\
    \bottomrule
    \end{tabular}}
  \caption{BLEU comparison between \mname and the baselines on zero-shot any-to-English language pairs. The Avg. column is the average BLEU over all zero-shot language pairs. The best BLEU score is bold and underlined.}
  \label{tab:task-results}
\end{table*} 
\paragraph{Training and evaluation} The Adam optimizer \cite{adam} with $\beta_1=0.9$ and $\beta_2=0.98$ is used for training. We use label smoothing with value 0.1. The learning rate is 0.0005 and warmup step is 4000 at the first stage. For the second stage, we set the learning rate as 0.0001 and do not use warmup. All the drop-out probabilities are set to 0.3. We use eight GPUs and the batch size is set as 4096 tokens per GPU. Maximum updates number is 200k for the first stage and 30k for the second stage. We use beam search (beam size is 5) and do not tune length penalty. We evaluate the results with sacrebleu\footnote{BLEU+case.mixed+numrefs.1+smooth.exp+tok.13a\\+version.1.5.0}. If not specified, the best checkpoint is selected by zero-shot cross-lingual transfer performance on the validation set for all experiments. We refer the reader to Section~\ref{sec:app-model-train} in Appendix for more training details.
\begin{table*}[t]
  \centering 
  \resizebox{1.0\textwidth}{!}{
  \begin{tabular}{l|ccc|r|rrrrrrrrrrr}
  \toprule
  ID  &  TwoStage   &  BigDec     &  Resdrop   & De    &  Nl    & Es    & It    &   Ro   & Fi     & Lv    & Et     &  Hi   & Ne   & Zh   & Avg.   \\ \midrule
  (1) &  $\times$  & $\times$  &  $\times$     & 23.7  & 32.5   & 20.2  & 25.7  &  20.3  & 12.3   & 9.2   & 10.9   & 7.8   & 5.4  & 6.3  & 15.1  \\  \midrule
  (2) &  \checkmark &   $\times$  &  $\times$  & 26.3  & 36.4   & 19.0  & 24.4  &  21.6  & 15.3   & 10.9  & 12.8   & 7.1   & 4.6  & 6.8  & 15.9   \\  
  (3) &   $\times$  &  \checkmark &  $\times$  & 26.4  & 32.7   & 20.7  & 26.1  &  21.5  & 13.7   & 9.4   & 11.5   & 7.7   & 5.0  & 6.8  & 15.5   \\      
    \midrule
  (4) &  \checkmark &  \checkmark &  $\times$  & \tub{27.3}  & 37.8   & 22.4  & 31.0  &  23.3  & 15.1   & 11.5  & 13.9   & 8.3   & 5.8  & 7.6  & 17.7   \\   
  (5) &  \checkmark &  $\times$ & \checkmark   & 25.7  & 36.4   & 19.4  & 25.8  &  22.5  & \tub{15.9}   & 11.1  & 13.3   & 7.6   & 5.2  & \tub{8.1}  & 16.5   \\   \midrule
  (6) &  \checkmark &  \checkmark & \checkmark & 26.4  & \tub{38.6}   & \tub{22.9}  & \tub{32.0}  &  \tub{23.9}  & 15.8   & \tub{12.0}  & \tub{14.5}   & \tub{8.6}   & \tub{6.1}  & \tub{8.1}  & \tub{18.3}   \\  
  \bottomrule
  \end{tabular}}
  \caption{Ablation study of the \mname trained on Europarl De-En. We compare models with different combinations of the second training stage (TwoStage),  the capacity-enhanced decoder (BigDec), and the position disentangled encoder (Resdrop). If using Resdrop, TwoStage is required because Resdrop is applied at the second training stage. Note that the model of ID (1) corresponds to the Strategy (7) in Table~\ref{tab:init-results} and ID (6) corresponds to \mname. The best BLEU score is bold and underlined.} 
  \label{tab:ablation}
 \end{table*}  

\paragraph{Baselines} We compare our model with vanilla Transformer and five conventional methods to apply pretrained Transformer encoder on NMT task. The pretrained encoders in these methods are replaced with XLM-R base for fair comparison.

$\bullet$ \noindent Vanilla Transformer. The encoder is with the same size of XLM-R base, the decoder uses the size of BaseDec. All model parameters are randomly initialized.

$\bullet$ \noindent +XLM-R fine-tune encoder~\cite{conneau2019cross}. The encoder is initialized with XLM-R. All parameters are trained.

$\bullet$ \noindent +XLM-R fine-tune all~\cite{conneau2019cross}. All parameters except those of cross attention module are initialized with XLM-R and directly fine-tuned. 

$\bullet$ \noindent +XLM-R as encoder embedding~\cite{zhu20incorporating}. The XLM-R output is leveraged as the encoder input of the NMT. The XLM-R model is fixed during training.

$\bullet$ \noindent +Recycle XLM-R for NMT~\cite{imamura2019recycling}. The method initializes the encoder with XLM-R and only trains decoder at the first step. Then all are trained at the second step.

$\bullet$ \noindent XLM-R fused model~\cite{zhu20incorporating}. The XLM-R output is fused into encoder and decoder separately with attention mechanism. The encoder embedding is initialized from XLM-R to facilitate transfer. The parameters of XLM-R are frozen during training.

\begin{table*}[th]
  \centering 
  \resizebox{1.0\textwidth}{!}{
    \begin{tabular}{ll|rr|rrr|rrr|rrrr|rrr|r}
    \toprule 
    \multirow{2}{*}{Model}  & \multirow{2}{*}{\# Sents} & \multicolumn{2}{c|}{German} & \multicolumn{3}{c|}{Romance} & \multicolumn{3}{c|}{Uralic} & \multicolumn{4}{c|}{Indo-Aryan} & \multicolumn{3}{c|}{East Asian} &  \multirow{2}{*}{Avg.} \\ 
              &       &  De   & Nl    & Es    & Ro    & It    & Fi    & Lv    & Et    & Hi    & Ne    & Si    & Gu    & Zh    & Ja    & Ko    &       \\ \midrule 
    mBART     & 0.04B & 27.4  & 43.3  & 24.7  & 28.2  & 29.8  & 18.8  & 14.2  & 15.7  & 12.3  & 9.6   & 7.2   & 10.3  &  8.3  &  6.0  & 21.1  & 18.4     \\
    CRISS     & 1.8B  & 28.8  & 47.0  & \tub{32.2}  & \tub{35.4}  & 48.9  & 23.9  & 18.6  & 23.5  & \tub{23.1}  & \tub{14.7}  & \tub{14.4}  & \tub{19.0}  & 13.4  &  7.9  & 24.8  & 25.0  \\ 
    m2m-100   & 7.5B  & 28.0  & 48.5  & 30.0  & 34.1  & \tub{50.0}  & 24.9  &\tub{19.9}  & \tub{25.8}  & 21.9  &  3.7  & 10.6  &  0.4  & \tub{19.5}  & \tub{11.5}  & \tub{32.7}  & 24.1   \\ \midrule
    \mname    & 0.04B & \tub{33.8}  & \tub{54.7}  & 30.1  & 33.9  & 43.0  & \tub{26.3}  & 17.7  & 25.7  & 17.5  & 14.4  & 12.2  & 17.3  & 13.4  & 10.7  & 31.2  & \tub{25.5}    \\ 
    \bottomrule
    \end{tabular}}
  \caption{Comparison with mBART, CRISS and m2m-100 on any-to-English test sets. Here we implement \mname with \mname large. mBART follows the original paper~\cite{mbart} for fine-tuning. `\# Sents' is the number of sentences in the NMT training set. The best BLEU score is bold and underlined. `Avg.' is the average BLEU across all language pairs.}
  \label{tab:compare-with-m2m}
\end{table*}

\subsection{Results} \label{sec:task-res}
The results of the empirical exploration in the Section~\ref{sec:empirical} are shown in Table~\ref{tab:init-results}. Since Strategy (8)--(9) use a larger decoder than the rest ones, we add Strategy (10) whose decoder size is the same as Strategy (8)--(9) for fair comparison. Overall, we observe that it is best to use a big decoder and initialize the decoder embedding and all encoder parameters with XLM-R, and to train the decoder layers from scratch (Strategy (10)). 

To verify the effect of a capacity enhanced decoder in the \tname task, we train vanilla Transformer with the same size of Strategy (7) (with BaseDec) and Strategy (10) (with BigDec) using the same training corpus.\footnote{We use De-En validation dataset this time.} The vanilla Transformer model with BaseDec and BigDec obtains a BLEU score of 23.5 and 22.9 on the De-En test set, respectively. The big decoder improves the performance of \mnamec, but fails to improve that of vanilla Transformer. This proves the effectiveness of BigDec to improve the zero-shot translation performance of our model. 

Table~\ref{tab:task-results} illustrates the performance of the proposed \mname comparing with the baselines. \mname gets 18.3 average BLEU and improves over the best baseline by 5.4 average BLEU, showing that \mname successfully learns to translate while preserving the cross-lingual transferability of XLM-R. For all language pairs, \mname obtains better transferring scores. In contrast, vanilla Transformer can hardly transfer and the other baselines do not well transfer to the distant languages. In addition to zero-shot performance, \mname also achieves the best result on De-En test set. Note that the best checkpoint is selected with zero-shot validation set for all methods.

Previous work \cite{xlmr,hu20xtreme} mainly uses XLM-R for cross-lingual transfer on NLU tasks. The experiments demonstrate that XLM-R can be also utilized for zero-shot neural machine translation if it is fine-tuned properly. We leave the exploration of cross-lingual transfer using XLM-R for other NLG tasks as the future work.

\subsection{Ablation Study} 

We conduct an ablation study with the proposed \mname on the Europarl De-En training set, as shown in Table~\ref{tab:ablation}. Overall, \mname obtains the best zero-shot translation results, demonstrating the importance of all three components. From the results of (1) to (3), TwoStage and BigDec along improve the zero-shot translation performance by 0.8 and 0.4 average BLEU over (1), respectively. However, combining them together brings a significant improvement of 2.6 average BLEU over (1). This indicates that TwoStage and BigDec are complementary to each other, thus it is important to use them together. The results of (6)$\rightarrow$(5) confirms our claim: without using BigDec, the performance of \mname drops by 1.8 average BLEU. We also observe that the supervised task (De-En) improves with TwoStage and BigDec (from results of (1) to (4)) while degrades with Resdrop (see results of (2)$\rightarrow$(5) and (4)$\rightarrow$(6)). This is expected since Resdrop helps to build a more language-agnostic encoder. Although Resdrop degrades supervised performance, it improves zero-shot translation. The zero-shot performance is related with both supervised performance and model transferability. By either enhancing the supervised performance (with TwoStage and BigDec) or the model transferability (with Resdrop), the overall performance of zero-shot translation can be improved.  

\begin{table*}[thb]
  \centering 
  \resizebox{1.0\textwidth}{!}{
    \begin{tabular}{ll|rr|rrr|rrr|rrrr|rrr|r}
    \toprule 
    & \multirow{2}{*}{Train set }  & \multicolumn{2}{c|}{German} & \multicolumn{3}{c|}{Romance} & \multicolumn{3}{c|}{Uralic} & \multicolumn{4}{c|}{Indo-Aryan} & \multicolumn{3}{c|}{East Asian} &  \multirow{2}{*}{Avg.} \\ 
          &                 &   De  & Nl    & Es    & Ro    & It    & Fi    & Lv    & Et      & Hi    & Ne   & Si   & Gu   & Zh   & Ja   & Ko    &  \\ \midrule
   \mtrv  & WMT19 De-En     & 33.7  & 3.0  & 3.6   & 3.4  & 1.7  & 1.6   & 1.2  & 1.8  & 0.1  & 0.1   & 0.2  & 0.2  & 0.3  & 0.7  & 0.3  & 3.5    \\ 
          & CCAligned Es-En &  6.8  & 5.5  & 32.5  & 6.4  & 17.3 & 2.0   & 1.7  & 2.5  & 0.3  & 0.1   & 0.2  & 0.2  & 0.8  & 0.8  & 0.4  & 5.2  \\ 
          & WMT19 Fi-En     &  1.3  & 0.7  &  1.3  & 1.8  & 0.6  & 21.7  & 0.6  & 1.7  & 0.2  & 0.1   & 0.1  & 0.2  & 0.2  & 0.4  & 0.3  & 2.1    \\ 
          & WAT21 Hi-En     &  0.6  & 0.9  &  0.2  & 0.5  & 0.6  & 0.4   & 0.3  & 0.5  & 21.5 & 3.6   & 0.1  & 0.2  & 0.1  & 0.1  & 0.3  & 2.0    \\ 
          & WMT18 Zh-En     &  0.2  & 0.2  & 0.3   & 0.3  & 0.2  & 0.3   & 0.1  & 0.3  & 0.1  & 0.1   & 0    & 0.1  & 22.3 & 0.3  & 0.1  & 1.7  \\ \midrule
   \mtrs  & WMT19 De-En     &  \tub{31.8}  & \tub{47.7}  & 24.6  & 18.9  & 36.5  & \tub{28.4}  & \tub{14.6}  & \tub{18.1}  & 9.5  & 6.4  & 7.2  & 9.5   & 9.6   & 6.8  & 22.4  & \tub{19.5}    \\ 
          & CCAligned Es-En  &  19.9  &  38.2 &  \tub{33.0} & \tub{30.9} & \tub{47.0} & 15.2 & 11.5 & 12.7 &  6.9 & 4.2 &  3.4   & 5.6 & 7.6  &  4.0   & 12.8  &  16.9 \\
          & WMT19 Fi-En     &  18.9  & 28.4  & 19.5  & 21.1  & 25.1  & 22.8  & 11.7  & 16.7  & 7.5  & 6.1   & 6.1  & 7.3   & 8.2   & 5.1  & 15.2  & 14.6    \\ 
          & WAT21 Hi-En     &  19.0  & 38.0  & 20.1  & 20.7  & 34.3  & 15.2  & 11.5  & 14.6  & \tub{24.3}  & \tub{16.7} &  \tub{9.6}  & \tub{17.8} &  8.3   & 5.9  & \tub{23.9}  & 18.7    \\ 
          & WMT18 Zh-En     &  20.0  & 31.8  & 21.2  & 21.8  & 28.2  & 15.1  & 11.4  & 13.4  & 10.6 & 7.4 &   8.1  & 8.2   & \tub{19.9}   & \tub{7.1}  & 20.2  &  16.3    \\ 
    \bottomrule
    \end{tabular}}
  \caption{The BLEU results of \mname with training data of different language pairs. The best BLEU of each test set with \mname model is bold and underlined. `Avg.' is the average BLEU across all language pairs.}
  \label{tab:dif-fam-results}
\end{table*}

\section{Analysis} \label{sec:expo}

\paragraph{Comparison with multilingual NMT} \label{sec:compare-m2m}
In this part, we compare \mname with mBART \cite{mbart}, CRISS \cite{tran2020criss} and m2m-100 \cite{fan2020m2m} on any-to-English test sets. mBART is a strong pretrained multilingual encoder-decoder based Transformer explicitly designed for NMT. We follow their setting and directly fine-tune all model parameters on WMT19 De-En training set. CRISS and m2m-100 are the state-of-the-art unsupervised and supervised multilingual NMT models, respectively. The CRISS model is initialized with the mBART model and iteratively fine-tuned on 1.8 billion sentences covering 90 language pairs. m2m-100 is trained with 7.5 billion parallel sentences across 2200 translation directions. The results of CRISS and m2m-100 are listed as reference, because CRISS and m2m-100 are many-to-many NMT models whose performance may degrade due to the competitions among different target languages \cite{aharoni2019massively,zhang2020improving}, while \mname is a many-to-one NMT model. The official m2m-100 model has three sizes: small (418M parameters), base (1.2B parameters) and large (12B parameters). The results of m2m-100 (small) model are reported.

To compare with these models, we train a many-to-one \mname large model with WMT19 German-English training data, which only consists of 41 million sentences pairs. It only requires a pretrained XLM-R large model and do not contain any data in other languages. We remove the residual connection after the self-attention sublayer of the 23-th (penultimate) encoder layer. The dataset and model configuration details are in Table~\ref{tab:traval-dataset} and~\ref{tab:app-dec-config} in the appendix. 

From the results in Table~\ref{tab:compare-with-m2m}, the \mname large model is significantly better than mBART and slightly better than CRISS and m2m-100. The averaged BLEU across all languages is 7.1, 0.5 and 1.4 higher than mBART, CRISS and m2m-100\footnote{The 1.2B m2m-100 model is larger than our model (737M parameters) and gets 2.2 more average BLEU than \mnamec.}, respectively.  The \mname model has larger model size, nevertheless, the results of \mname are impressive given that \mname does not use any monolingual or parallel texts except German-English training data. The performance gain over mBART shows that with proper fine-tuning strategy, the pretrained multilingual encoder has better cross-lingual transfer ability on NMT tasks. In addition, with large-scale German-English parallel data, the \mname model transfers well to distant resource-poor languages like Ne and Si, which indicates a promising approach to translate resource-poor languages. The \mname performance might be further improved with the data of more languages pairs. We leave this as future work.
\begin{table}[t]
  \centering 
  \resizebox{0.85\columnwidth}{!}{
    \begin{tabular}{llrr}
    \toprule 
    Train set        & \# Sents  & Vanilla  & \mname \\ \midrule
    Europarl De-En   & 1.9M     & 23.1  & 26.4   \\ 
    WAT21 Hi-En      & 3.5M     & 26.1  & 24.3    \\
    WMT16 De-En      & 4.5M     & 30.9  & 31.2   \\ 
    WMT19 Fi-En      & 4.8M     & 22.5  & 22.8    \\
    CCAligned Es-En  & 20M      & 37.5  & 33.0    \\
    WMT18 Zh-En      & 23M      & 22.3  & 19.9    \\
    WMT19 De-En      & 41M      & 33.7  & 31.8   \\ 
    \bottomrule   
    \end{tabular}}
  \caption{Comparison with vanilla Transformer on the supervised translation direction. The `\# Sents' column is the number of sentence pairs of the dataset.}
  \label{tab:vanilla-results}
\end{table}

\begin{table*}[t]
  \centering 
  \resizebox{1.0\textwidth}{!}{
    \begin{tabular}{ll|rr|rrr|rrr|rrrr|rrr|r}
    \toprule 
    \multirow{2}{*}{Train set}  & \multirow{2}{*}{\# Sents} & \multicolumn{2}{c|}{German} & \multicolumn{3}{c|}{Romance} & \multicolumn{3}{c|}{Uralic} & \multicolumn{4}{c|}{Indo-Aryan} & \multicolumn{3}{c|}{East Asian} &  \multirow{2}{*}{Avg.} \\ 
              &          &  De   & Nl    & Es    & Ro    & It    & Fi    & Lv    & Et    & Hi   & Ne   & Si   & Gu   & Zh   & Ja    & Ko    &       \\ \midrule 
    Europarl-v7  & 1.9M    & 26.4  & 38.6  & 22.9  & 23.9  & 32.0  & 15.8  & 12.0  & 14.5  & 8.6  & 6.1  & 5.6  & 7.5  & 8.1  &  4.7  & 15.1  & 16.1    \\
    WMT19        & 41M   & 31.8  & 47.7  & 24.6  & 18.9  & 36.5  & 28.4  & 14.6  & 18.1  & 9.5  & 6.4  & 7.2  & 9.5  & 9.6  &  6.8  & 22.4  & 19.5  \\ 
    \bottomrule
    \end{tabular}}
  \caption{The BLEU results of \mname with training data of different sizes for any-to-English translation. `\# Sents' is the number of parallel sentences in the training set. ‘Avg.’ is the average BLEU across all language pairs.}
  \label{tab:dif-size-results}
\end{table*}
\paragraph{Language transfer v.s. language distance} \label{sec:dif-fam}
In this part, we explore the relationship between the cross-lingual transfer performance and the language distance. We train the \mname models on different supervised language pairs including De-En, Es-En, Fi-En, Hi-En and Zh-En, and then directly apply them to all test sets, as seen in Table~\ref{tab:dif-fam-results}.\footnote{The details of the datasets are in the appendix.} We observe that the cross-lingual transfer generally works better when the \mname model is trained on source languages in the same language family. The performance on Ko-En is one exception, where Hi-En achieves the best transfer performance. We also notice that the vocabulary overlapping (even character overlapping) between Hindi and Korean is low, showing that significant vocabulary sharing is not a requirement for effective transfer. When trained on 3.5 million Hi-En sentence pairs, \mname obtains promising results on the Ne-En and Si-En translation, with a BLEU score of 16.7 and 9.6, respectively. As comparison, The vanilla Transformer supervised with FLoRes training set only receives 14.5 and 7.2 BLEU score~\cite{mbart} on the same test sets. Therefore, another approach to translate resource-poor languages is to train \mname on similar high-resource language pairs.

As a comparison, we train vanilla Transformer configured as Transformer big\footnote{i.e. `transformer\_wmt\_en\_de\_big' configuration in the \texttt{fairseq} toolkit.} without MPE initialization with the same training sets and validation sets. The poor zero-shot cross-lingual performance of vanilla Transformer indicates that the XLM-R initialized encoder is essential and can produce language-agnostic representations. 

\paragraph{Performance on the supervised language pair} \label{sec:vanilla-performance}
To study whether the \mname model gains the cross-lingual transfer ability at the cost of performance degradation on the supervised language pair, we compare the vanilla Transformer big model\footnote{The validation dataset of the supervised language pair is used.} and \mname model on the supervised translation task, as shown in Table~\ref{tab:vanilla-results}. The performance of \mname is lower than that of vanilla Transformer when more than 20M parallel sentences are available, but it gets better performance with fewer parallel sentences. The Hindi-to-English is an exception where \mname has lower BLEU. When large amount of bi-text data is given, the \mname model size is expected to be increased to fully digest the bi-text. For example, if we replace \mname with \mname large and train \mname large on WMT19 De-En, we get 33.8 BLEU on De-En test set (see Table~\ref{tab:compare-with-m2m}), which is comparable of 33.7 BLEU obtained by vanilla Transformer. 

\paragraph{Performance v.s. training corpus size } \label{sec:dif-size}
To examine the relationship between cross-lingual transfer ability and training data size, we compare the zero-shot BLEU scores of \mname models trained on Europarl De-En and WMT19 De-En. The results are shown in Table~\ref{tab:dif-size-results}. It shows that increasing training data size can consistently improve the zero-shot translation performance. For instance, \mname trained with WMT19 improves over \mname trained with Europarl-v7 by 3.4 average BLEU.

\paragraph{Performance with other target language} \label{sec:ende}
To build many-to-one NMT model with other target language, we train two \mname models on WMT16 En-De and WMT19 En-De, respectively. We use Fi-De as validation language pair and Fr/Cs/Ru/Nl-De as test language pairs. From the results shown in Table~\ref{tab:ende-results}, \mname can obtain reasonable transferring scores to unseen source languages when target language is not English. Again, the results confirm that the cross-lingual transfer ability improves with larger training data.

\begin{table}[t]
  \centering 
  \resizebox{1.0\columnwidth}{!}{
    \begin{tabular}{l|r|rrrrrr}
    \toprule
    Train set   & En-De  & Fr-De  & Cs-De  & Ru-De & Nl-De  & Avg.    \\ \midrule
    WMT16    & 25.7   & 18.5    & 14.4  & 29.0  & 39.0  & 25.2    \\ 
    WMT19    & 26.7   & 20.1    & 15.6  & 31.4  & 42.3  & 27.4    \\ 
    \bottomrule
    \end{tabular}}
  \caption{The BLEU results of \mname for any-to-German translation. `Avg.' denotes the average BLEU across all zero-shot language pairs.}
  \label{tab:ende-results}
\end{table} 

\section{Related Work}
\paragraph{Zero-shot cross-lingual transfer learning} Multilingual pretrained models, such as mBERT~\cite{mbert}, XLM-R~\cite{xlmr}, mBART~\cite{mbart}, and mT5~\cite{xue2020mt5}, have achieved success on zero-shot cross-lingual transfer for various NLP tasks. The models are pretrained on large-scale multilingual corpora with a shared vocabulary. After pretrained, it is fine-tuned on labeled data of downstream tasks in one language and directly tested in other languages in a zero-shot manner. While multilingual pretrained models with encoder-decoder-based architecture \cite{mbart,chi2020cross} work well on cross-lingual transfer for NLG tasks, multilingual pretrained encoders~\cite{mbert,conneau2019cross,xlmr} are mainly applied to cross-lingual NLU tasks \cite{hu20xtreme}. In this work, we explore how to fine-tune an off-the-shelf multilingual pretrained encoder for zero-shot cross-lingual transfer in neural machine translation, a typical NLG task. 

\paragraph{Pretrained models for NMT} Some previous works~\cite{imamura2019recycling,conneau2019cross,yang20towards,weng20acquiring,ma2020xlmt,zhu20incorporating} explore methods to integrate pretrained language encoders into the NMT model to improve supervised translation performance. For instance, \citet{zhu20incorporating} propose BERT-fused model, in which they first use BERT to extract representations for an input sentence, and then fuses the representations into both the encoder and decoder via the attention mechanism. Another line of works~\cite{mbart,song2019mass,lin2020pre} propose novel encoder-decoder-based multilingual pretrained language models and fine-tune such models for NMT. For example, \citet{mbart} propose mBART, an encoder-decoder-based Transformer explicitly designs for NMT and demonstrate that mBART can be fine-tuned for supervised and zero-shot NMT. Different from them, we leverage MPE for zero-shot translation instead of supervised translation. Among the previous works, \citet{wei2020learning} is the most similar with ours. They fine-tune their MPE on NMT with a two-stage strategy. However, their work focuses on improving the MPE for a more universal representation across languages and lacks in-depth study of cross-lingual NMT. In contrast, we aim at leveraging an MPE for machine translation while preserving its ability of cross-lingual transfer. 

\section{Conclusion}
In this paper, we focus on the zero-shot cross-lingual NMT transfer~(\tnamec) task which aims at leveraging an MPE for machine translation while preserving its ability of cross-lingual transfer. In this task, only a multilingual pretrained encoder such as XLM-R and one parallel dataset such as German-English are available. We propose \mname for this task, which enables zero-shot cross-lingual transfer for NMT by making full use of the labelled data and enhancing the transferability of XLM-R. Extensive experiments demonstrate the effectiveness of \mnamec. In particular, \mname outperforms mBART, a pretrained encoder-decoder-based model explicitly designed for NMT. It also gets better performance than CRISS and m2m-100, two strong multilingual NMT models, on 15 any-to-English test sets with less training data and training computation cost.

\section*{Acknowledgements}
This project was supported by National Natural Science Foundation of China (No. 62106138) and Shanghai Sailing Program (No. 21YF1412100). Wenping Wang and Jia Pan acknowledge the support from Centre for Transformative Garment Production. We thank the anonymous reviewers for their insightful feedbacks on this work.

\bibliography{anthology,custom}

\begin{thebibliography}{32}
\expandafter\ifx\csname natexlab\endcsname\relax\def\natexlab#1{#1}\fi

\bibitem[{Aharoni et~al.(2019)Aharoni, Johnson, and
  Firat}]{aharoni2019massively}
Roee Aharoni, Melvin Johnson, and Orhan Firat. 2019.
\newblock Massively multilingual neural machine translation.
\newblock In \emph{Proceedings of NAACL}, pages 3874--3884.

\bibitem[{Chen et~al.(2017)Chen, Liu, Cheng, and Li}]{chen2017teacher}
Yun Chen, Yang Liu, Yong Cheng, and Victor~OK Li. 2017.
\newblock A teacher-student framework for zero-resource neural machine
  translation.
\newblock In \emph{Proceedings of ACL}, pages 1925--1935.

\bibitem[{Chen et~al.(2018)Chen, Liu, and Li}]{chen2018zero}
Yun Chen, Yang Liu, and Victor~OK Li. 2018.
\newblock Zero-resource neural machine translation with multi-agent
  communication game.
\newblock In \emph{Thirty-Second AAAI Conference on Artificial Intelligence}.

\bibitem[{Chi et~al.(2020)Chi, Dong, Wei, Wang, Mao, and Huang}]{chi2020cross}
Zewen Chi, Li~Dong, Furu Wei, Wenhui Wang, Xian-Ling Mao, and Heyan Huang.
  2020.
\newblock Cross-lingual natural language generation via pre-training.
\newblock In \emph{Proceedings of the AAAI Conference on Artificial
  Intelligence}, volume~34, pages 7570--7577.

\bibitem[{Conneau et~al.(2020)Conneau, Khandelwal, Goyal, Chaudhary, Wenzek,
  Guzm{\'a}n, Grave, Ott, Zettlemoyer, and Stoyanov}]{xlmr}
Alexis Conneau, Kartikay Khandelwal, Naman Goyal, Vishrav Chaudhary, Guillaume
  Wenzek, Francisco Guzm{\'a}n, Edouard Grave, Myle Ott, Luke Zettlemoyer, and
  Veselin Stoyanov. 2020.
\newblock Unsupervised cross-lingual representation learning at scale.
\newblock In \emph{Proceedings of ACL}, pages 8440--8451, Online.

\bibitem[{Conneau and Lample(2019)}]{conneau2019cross}
Alexis Conneau and Guillaume Lample. 2019.
\newblock Cross-lingual language model pretraining.
\newblock In \emph{Advances in Neural Information Processing Systems}, pages
  7059--7069.

\bibitem[{Currey et~al.(2020)Currey, Mathur, and Dinu}]{currey2020distilling}
Anna Currey, Prashant Mathur, and Georgiana Dinu. 2020.
\newblock Distilling multiple domains for neural machine translation.
\newblock In \emph{Proceedings of EMNLP}, pages 4500--4511.

\bibitem[{Devlin et~al.(2019)Devlin, Chang, Lee, and Toutanova}]{bert}
Jacob Devlin, Ming-Wei Chang, Kenton Lee, and Kristina Toutanova. 2019.
\newblock {BERT}: Pre-training of deep bidirectional transformers for language
  understanding.
\newblock In \emph{Proceedings of NAACL}, pages 4171--4186.

\bibitem[{Fan et~al.(2020)Fan, Bhosale, Schwenk, Ma, El-Kishky, Goyal, Baines,
  Celebi, Wenzek, Chaudhary, Goyal, Birch, Liptchinsky, Edunov, Grave, Auli,
  and Joulin}]{fan2020m2m}
Angela Fan, Shruti Bhosale, Holger Schwenk, Zhiyi Ma, Ahmed El-Kishky,
  Siddharth Goyal, Mandeep Baines, Onur Celebi, Guillaume Wenzek, Vishrav
  Chaudhary, Naman Goyal, Tom Birch, Vitaliy Liptchinsky, Sergey Edunov,
  Edouard Grave, Michael Auli, and Armand Joulin. 2020.
\newblock Beyond english-centric multilingual machine translation.
\newblock \emph{arXiv preprint arXiv:2010.11125}.

\bibitem[{Gu et~al.(2018)Gu, Wang, Chen, Li, and Cho}]{gu2018maml}
Jiatao Gu, Yong Wang, Yun Chen, Victor O.~K. Li, and Kyunghyun Cho. 2018.
\newblock Meta-learning for low-resource neural machine translation.
\newblock In \emph{Proceedings of EMNLP}, pages 3622--3631.

\bibitem[{Hu et~al.(2020)Hu, Ruder, Siddhant, Neubig, Firat, and
  Johnson}]{hu20xtreme}
Junjie Hu, Sebastian Ruder, Aditya Siddhant, Graham Neubig, Orhan Firat, and
  Melvin Johnson. 2020.
\newblock {XTREME}: A massively multilingual multi-task benchmark for
  evaluating cross-lingual generalisation.
\newblock In \emph{Proceedings of the 37th International Conference on Machine
  Learning}, volume 119, pages 4411--4421.

\bibitem[{Imamura and Sumita(2019)}]{imamura2019recycling}
Kenji Imamura and Eiichiro Sumita. 2019.
\newblock Recycling a pre-trained {BERT} encoder for neural machine
  translation.
\newblock In \emph{Proceedings of the 3rd Workshop on Neural Generation and
  Translation}, pages 23--31.

\bibitem[{Johnson et~al.(2017)Johnson, Schuster, Le, Krikun, Wu, Chen, Thorat,
  Vi{\'e}gas, Wattenberg, Corrado et~al.}]{johnson2017google}
Melvin Johnson, Mike Schuster, Quoc~V Le, Maxim Krikun, Yonghui Wu, Zhifeng
  Chen, Nikhil Thorat, Fernanda Vi{\'e}gas, Martin Wattenberg, Greg Corrado,
  et~al. 2017.
\newblock Google’s multilingual neural machine translation system: Enabling
  zero-shot translation.
\newblock \emph{Transactions of the Association for Computational Linguistics},
  5:339--351.

\bibitem[{Kingma and Ba(2015)}]{adam}
Diederik~P. Kingma and Jimmy Ba. 2015.
\newblock Adam: A method for stochastic optimization.
\newblock In \emph{Proceedings of ICLR}.

\bibitem[{Kudo(2018)}]{sentencepiece}
Taku Kudo. 2018.
\newblock Subword regularization: Improving neural network translation models
  with multiple subword candidates.
\newblock In \emph{Proceedings of ACL}, pages 66--75.

\bibitem[{Lample et~al.(2018)Lample, Ott, Conneau, Denoyer, and
  Ranzato}]{lample2018phrase}
Guillaume Lample, Myle Ott, Alexis Conneau, Ludovic Denoyer, and Marc'Aurelio
  Ranzato. 2018.
\newblock Phrase-based \& neural unsupervised machine translation.
\newblock In \emph{Proceedings of EMNLP}.

\bibitem[{Lin et~al.(2020)Lin, Pan, Wang, Qiu, Feng, Zhou, and Li}]{lin2020pre}
Zehui Lin, Xiao Pan, Mingxuan Wang, Xipeng Qiu, Jiangtao Feng, Hao Zhou, and
  Lei Li. 2020.
\newblock Pre-training multilingual neural machine translation by leveraging
  alignment information.
\newblock In \emph{Proceedings of the 2020 Conference on Empirical Methods in
  Natural Language Processing (EMNLP)}, pages 2649--2663.

\bibitem[{Liu et~al.(2021)Liu, Niehues, Cross, Guzm{\'a}n, and
  Li}]{liu2020improving}
Danni Liu, Jan Niehues, James Cross, Francisco Guzm{\'a}n, and Xian Li. 2021.
\newblock Improving zero-shot translation by disentangling positional
  information.
\newblock In \emph{Proceedings of ACL}, pages 1259--1273.

\bibitem[{Liu et~al.(2020)Liu, Gu, Goyal, Li, Edunov, Ghazvininejad, Lewis, and
  Zettlemoyer}]{mbart}
Yinhan Liu, Jiatao Gu, Naman Goyal, Xian Li, Sergey Edunov, Marjan
  Ghazvininejad, Mike Lewis, and Luke Zettlemoyer. 2020.
\newblock Multilingual denoising pre-training for neural machine translation.
\newblock \emph{arXiv preprint arXiv:2001.08210}.

\bibitem[{Ma et~al.(2020)Ma, Yang, Huang, Chi, Dong, Zhang, Awadalla, Muzio,
  Eriguchi, Singhal, Song, Menezes, and Wei}]{ma2020xlmt}
Shuming Ma, Jian Yang, Haoyang Huang, Zewen Chi, Li~Dong, Dongdong Zhang,
  Hany~Hassan Awadalla, Alexandre Muzio, Akiko Eriguchi, Saksham Singhal, Xia
  Song, Arul Menezes, and Furu Wei. 2020.
\newblock Xlm-t: Scaling up multilingual machine translation with pretrained
  cross-lingual transformer encoders.
\newblock \emph{arXiv preprint arXiv:2012.15547}.

\bibitem[{Ott et~al.(2019)Ott, Edunov, Baevski, Fan, Gross, Ng, Grangier, and
  Auli}]{ott2019fairseq}
Myle Ott, Sergey Edunov, Alexei Baevski, Angela Fan, Sam Gross, Nathan Ng,
  David Grangier, and Michael Auli. 2019.
\newblock fairseq: A fast, extensible toolkit for sequence modeling.
\newblock In \emph{Proceedings of NAACL}.

\bibitem[{Serra et~al.(2018)Serra, Suris, Miron, and
  Karatzoglou}]{serra18overcoming}
Joan Serra, Didac Suris, Marius Miron, and Alexandros Karatzoglou. 2018.
\newblock Overcoming catastrophic forgetting with hard attention to the task.
\newblock In \emph{Proceedings of ICML}, volume~80, pages 4548--4557.

\bibitem[{Song et~al.(2019)Song, Tan, Qin, Lu, and Liu}]{song2019mass}
Kaitao Song, Xu~Tan, Tao Qin, Jianfeng Lu, and Tie-Yan Liu. 2019.
\newblock Mass: Masked sequence to sequence pre-training for language
  generation.
\newblock In \emph{International Conference on Machine Learning}, pages
  5926--5936.

\bibitem[{Tran et~al.(2020)Tran, Tang, Li, and Gu}]{tran2020criss}
Chau Tran, Yuqing Tang, Xian Li, and Jiatao Gu. 2020.
\newblock Cross-lingual retrieval for iterative self-supervised training.
\newblock In \emph{Proceedings of NeurIPS}.

\bibitem[{Vaswani et~al.(2017)Vaswani, Shazeer, Parmar, Uszkoreit, Jones,
  Gomez, Kaiser, and Polosukhin}]{transformer}
Ashish Vaswani, Noam Shazeer, Niki Parmar, Jakob Uszkoreit, Llion Jones,
  Aidan~N Gomez, Lukasz Kaiser, and Illia Polosukhin. 2017.
\newblock Attention is all you need.
\newblock In \emph{NeurIPS}, pages 5998--6008.

\bibitem[{Wei et~al.(2021)Wei, Hu, Weng, Xing, Yu, and Luo}]{wei2020learning}
Xiangpeng Wei, Yue Hu, Rongxiang Weng, Luxi Xing, Heng Yu, and Weihua Luo.
  2021.
\newblock On learning universal representations across languages.
\newblock \emph{Proceedings of ICLR}.

\bibitem[{Weng et~al.(2020)Weng, Yu, Huang, Cheng, and Luo}]{weng20acquiring}
Rongxiang Weng, Heng Yu, Shujian Huang, Shanbo Cheng, and Weihua Luo. 2020.
\newblock Acquiring knowledge from pre-trained model to neural machine
  translation.
\newblock In \emph{Proceedings of AAAI}, pages 9266--9273.

\bibitem[{Wu and Dredze(2019)}]{mbert}
Shijie Wu and Mark Dredze. 2019.
\newblock Beto, bentz, becas: The surprising cross-lingual effectiveness of
  bert.
\newblock In \emph{Proceedings of EMNLP-IJCNLP}, pages 833--844.

\bibitem[{Xue et~al.(2021)Xue, Constant, Roberts, Kale, Al-Rfou, Siddhant,
  Barua, and Raffel}]{xue2020mt5}
Linting Xue, Noah Constant, Adam Roberts, Mihir Kale, Rami Al-Rfou, Aditya
  Siddhant, Aditya Barua, and Colin Raffel. 2021.
\newblock m{T}5: A massively multilingual pre-trained text-to-text transformer.
\newblock In \emph{Proceedings of NAACL}, pages 483--498.

\bibitem[{Yang et~al.(2020)Yang, Wang, Zhou, Zhao, Zhang, Yu, and
  Li}]{yang20towards}
Jiacheng Yang, Mingxuan Wang, Hao Zhou, Chengqi Zhao, Weinan Zhang, Yong Yu,
  and Lei Li. 2020.
\newblock Towards making the most of bert in neural machine translation.
\newblock In \emph{Proceedings of AAAI}, pages 9378--9385.

\bibitem[{Zhang et~al.(2020)Zhang, Williams, Titov, and
  Sennrich}]{zhang2020improving}
Biao Zhang, Philip Williams, Ivan Titov, and Rico Sennrich. 2020.
\newblock Improving massively multilingual neural machine translation and
  zero-shot translation.
\newblock In \emph{Proceedings of ACL}, pages 1628--1639.

\bibitem[{Zhu et~al.(2020)Zhu, Xia, Wu, He, Qin, Zhou, Li, and
  Liu}]{zhu20incorporating}
Jinhua Zhu, Yingce Xia, Lijun Wu, Di~He, Tao Qin, Wengang Zhou, Houqiang Li,
  and Tieyan Liu. 2020.
\newblock Incorporating bert into neural machine translation.
\newblock In \emph{Proceedings of ICLR}.

\end{thebibliography}
\bibliographystyle{acl_natbib}

\appendix
\section{Dataset} \label{sec:app-data}
The dataset is from WMT translation task, CCAligned corpus\footnote{\url{http://www.statmt.org/cc-aligned/}},  WAT21 translation task\footnote{\url{http://lotus.kuee.kyoto-u.ac.jp/WAT/indic-multilingual/indic_wat_2021.tar.gz}}, Flores test set\footnote{\url{https://github.com/facebookresearch/flores/raw/master/data/flores_test_sets.tgz}} and Tatoeba test sets\footnote{\url{https://object.pouta.csc.fi/Tatoeba-Challenge/test-v2020-07-28.tar}}. We use the first 20M sentence pairs in the Es-En CCAligned corpus as training set. For experiments of Table~\ref{tab:dif-fam-results}, the validation set for De-En, Es-En and Fi-En are the concatnation of Fr-En and Cs-En validation set. We use Ta-En and Zh-En as the validation set for Hi-En and Zh-En, respectively. More details are in Table~\ref{tab:traval-dataset} to Table~\ref{tab:ende-dataset}. 

To be compatible with XLM-R model, all texts are tokenized with the same XLM-R sentencepiece \cite{sentencepiece} model. The $<$bos$>$ token is added at the beginning of each source sentence while $<$eos$>$ token is appended at the end when the NMT model initializes encoder with XLM-R. The source sentence length is limited within 512 tokens. 

\begin{table}[htbp]
  \centering 
  \resizebox{0.92\columnwidth}{!}{
    \begin{tabular}{llll}
    \toprule
    Type          &   Lang          & Source          & \# Sents  \\ \midrule
    Training set  &   De-En         & Europarl v7     & 1.9M        \\ 
    Training set  &   De-En         & WMT16           & 4.5M      \\
    Training set  &   De-En         & WMT19           & 41M      \\
    Training set  &   Es-En         & CCAligned       & 20M      \\ 
    Training set  &   Fi-En         & WMT19           & 4.8M      \\ 
    Training set  &   Hi-En         & WAT21           & 3.5M      \\ 
    Training set  &   Zh-En         & WMT18           & 23M      \\ \midrule
    Valid set     &   Cs-En         & Newstest 14     & 3003      \\ 
    Valid set     &   Es-En         & Newstest 10     & 2489      \\
    Valid set     &   Fi-En         & Newstest 19     & 1996      \\
    Valid set     &   Fr-En         & Newstest 14     & 3003      \\ 
    Valid set     &   Hi-En         & Newsdev 14      & 520      \\
    Valid set     &   Ta-En         & WAT21           & 2390      \\ 
    Valid set     &   Zh-En         & Newstest 17     & 2001      \\
    \bottomrule
    \end{tabular}}
  \caption{Training and valid set for any-to-English translation. The `\# Sents' column is the number of sentence pairs of the dataset.}
  \label{tab:traval-dataset}
\end{table} 

\begin{table}[ht]
  \centering 
  \resizebox{0.85\columnwidth}{!}{
    \begin{tabular}{ll|ll}
    \toprule
    Lang   & Source        & Lang    & Source          \\ \midrule
    De-En  & Newstest 14   & Ko-En  & Tatoeba           \\
    Es-En  & Newstest 13   & Lv-En   & Newstest 17     \\ 
    Et-En  & Newstest 18   & Ne-En   & Flores          \\ 
    Fi-En  & Newstest 16   & Nl-En   & Tatoeba         \\
    Gu-En  & Newstest 19   & Ro-En   & Newstest 16     \\ 
    Hi-En  & Newstest 14   & Si-En   & Flores          \\ 
    It-En  & Tatoeba       & Zh-En   & Newstest 18     \\ 
    Ja-En  & Newstest 20   &  & \\
    \bottomrule
    \end{tabular}}
  \caption{Test sets for any-to-English translation.}
  \label{tab:test-dataset}
\end{table}

\begin{table}[!t]
  \centering 
  \resizebox{0.75\columnwidth}{!}{
    \begin{tabular}{lll}
    \toprule
    Type          &   Lang          & Source            \\ \midrule
    Training set  &   En-De         & WMT16             \\  
    Training set  &   En-De         & WMT19             \\  
    Valid set     &   Fi-De         & Tatoeba          \\ \midrule
    Test set      &   Cs-De         & Newstest 19      \\
    Test set      &   En-De         & Newstest 14      \\
    Test set      &   Fr-De         & Newstest 19      \\ 
    Test set      &   Nl-De         & Tatoeba          \\ 
    Test set      &   Ru-De         & Tatoeba          \\ 
    \bottomrule
    \end{tabular}}
  \caption{Dataset used for English-to-German translation in Section~\ref{sec:ende}.}
  \label{tab:ende-dataset}
\end{table} 

\begin{table*}[!t]
  \centering 
  \resizebox{1.5\columnwidth}{!}{
    \begin{tabular}{llllllll}
    \toprule
    Model                & $H_d$   & $H^{ff}_{enc}$ & $L_{enc}$  &  $A_{enc}$ & $H^{ff}_{dec}$  & $L_{dec}$  &  $A_{dec}$     \\ \midrule
    Transformer base     &  512  &  2048  &  6   &  8  &  2048  &  6   &  8     \\
    Transformer big      & 1024  &  4096  &  6   & 16  &  4096  &  6   &  16       \\
    SixT small   & 768   &  3072  &  12  & 12  &  2048  &  6   &  8      \\
    SixT     & 768   &  3072  &  12  & 12  &  3072  &  12  &  12     \\
    SixT large           & 1024  &  4096  &  24  & 16  &  3072  &  12  &  16    \\
    \bottomrule
    \end{tabular}}
  \caption{Model configurations for different models. The `A' column is the number of attention heads.}
  \label{tab:app-dec-config}
\end{table*} 

\section{Model and Training Details} \label{sec:app-model-train}
The encoder of \mname is the same size of XLM-R model. We compare models with different decoder configurations in the paper, the details are in the Table~\ref{tab:app-dec-config}. For all models, the dimension of decoder hidden states equals that of encoder hidden states. The number of attention heads is set as 16 for the decoder of \mname large model, so that the dimension of hidden states can be divided by the number of attention heads. We use separate encoder and decoder embeddings. We tie the decoder input and output embeddings. The source vocabulary uses the same 250k vocabulary of XLM-R, while the target vocabulary is generated from the training corpus. All experiments are done with 8 GPUs. 

We compare \mname large with CRISS, m2m-100 and mBART in the Table~\ref{tab:compare-with-m2m}. We use the official model checkpoints of mBART\footnote{\url{https://github.com/pytorch/fairseq/blob/master/examples/mbart}} (611M parameters), CRISS\footnote{\url{https://github.com/pytorch/fairseq/tree/master/examples/criss}} (680M parameters) and m2m-100\footnote{\url{https://github.com/pytorch/fairseq/tree/master/examples/m2m_100}} (418M parameters). The training hyper-parameters of \mname large model are the same with that in Section~\ref{sec:setup}. 

\section{Language Code}
The information of the languages used in this paper is listed in the Table~\ref{tab:app-isocode}. 
\begin{table}[ht]
  \centering 
  \resizebox{0.65\columnwidth}{!}{
    \begin{tabular}{lll}
    \toprule
    ISO   & Language  & Family   \\  \midrule
    cs & Czech      & Slavic     \\        
    de & German     & Germanic   \\               
    en & English    & Germanic   \\               
    es & Spanish    & Romance    \\    
    et & Estonian   & Uralic     \\                    
    fi & Finnish    & Uralic     \\                     
    fr & French     & Romance    \\       
    gu & Gujarati   & Indo-Aryan \\                 
    hi & Hindi      & Indo-Aryan \\                  
    it & Italian    & Romance    \\     
    ja & Japanese   & Japonic      \\
    ko & Korean     & Koreanic     \\
    lv & Latvian    & Baltic       \\
    ne & Nepali     & Indo-Aryan   \\
    nl & Dutch      & Germanic     \\
    ro & Romanian   & Romance      \\
    ru & Russian    & Slavic       \\
    si & Sinhala    & Indo-Aryan   \\
    ta & Tamil      & Dravidian    \\
    zh & Chinese    & Chinese      \\     
    \bottomrule
    \end{tabular}}
  \caption{The information of the languages used in this paper.}
  \label{tab:app-isocode}
\end{table} 

\end{document}